\documentclass[preprint,12pt]{elsarticle}

\usepackage{amssymb}

\usepackage{lineno}

\usepackage{amsmath,amsfonts}
\usepackage{graphicx}
\usepackage{textcomp}

\usepackage{multirow}
\usepackage{colortbl}
\usepackage{caption}

\usepackage{algpseudocode}
\usepackage{setspace}
\usepackage{color,hyperref}
\usepackage{subcaption}
\usepackage{graphicx}
\usepackage{amssymb} 
\usepackage{epsfig}
\usepackage{amsmath}
\usepackage{amssymb}
\usepackage{colortbl}
\usepackage{soul}
\usepackage{multirow}
\usepackage{pifont}

\usepackage{array}
\newcolumntype{I}{!{\vrule width 1.2pt}}
\newlength\savedwidth
\newcommand\whline{\noalign{\global\savedwidth\arrayrulewidth
		\global\arrayrulewidth 1.25pt}%
	\hline
	\noalign{\global\arrayrulewidth\savedwidth}}

\journal{Name of journal}

\begin{document}

\begin{frontmatter}



\title{SCAR: Spatial-/Channel-wise Attention Regression Networks for Crowd Counting}


\author{Junyu Gao, Qi Wang, Yuan Yuan}

\address{School of Computer Science and Center for OPTical IMagery Analysis and Learning (OPTIMAL), \\Northwestern Polytechnical University, \\ Xi'an {\rm 710072}, P. R. China}

\begin{abstract}
 
Recently, crowd counting is a hot topic in crowd analysis. Many CNN-based counting algorithms attain good performance. However, these methods only focus on the local appearance features of crowd scenes but ignore the large-range pixel-wise contextual and crowd attention information. To remedy the above problems, in this paper, we introduce the \textbf{S}patial-/\textbf{C}hannel-wise \textbf{A}ttention Models into the traditional \textbf{R}egression CNN to estimate the density map, which is named as ``SCAR''. It consists of two modules, namely Spatial-wise Attention Model (SAM) and Channel-wise Attention Model (CAM). The former can encode the pixel-wise context of the entire image to more accurately predict density maps at the pixel level. The latter attempts to extract more discriminative features among different channels, which aids model to pay attention to the head region, the core of crowd scenes. Intuitively, CAM alleviates the mistaken estimation for background regions. Finally, two types of attention information and traditional CNN's feature maps are integrated by a concatenation operation. Furthermore, the extensive experiments are conducted on four popular datasets, Shanghai Tech Part A/B, GCC, and UCF\_CC\_50 Dataset. The results show that the proposed method achieves state-of-the-art results.

\end{abstract}


\end{frontmatter}



\section{Introduction}
\label{intro}
Crowd analysis is a popular task in computer vision \mbox{\cite{ZITOUNI2016139,LU2017213,WEI2019,CHEUNG20191}}, which focuses on understand the still or video crowd scenes at a high level. In the field of crowd analysis, crowd counting \cite{ZHANG2015151,li2018csrnet,WANG2019360,MA201991} is an essential branch, which focuses on predicting the number of people or estimating the density maps for crowd scenes. Accurate crowd counting is important to urban safety, public design, space management and so on. In this paper, we aim to the crowd counting for video surveillance and still crowd scenes. 

With the development of deep learning on computer vision, scene understanding \cite{lu2017hybrid,lu2017exploring,zhao2017hierarchical,zhao2019cam,8766103,8693661} achieves a remarkable progress. At present, CNN-based methods \cite{gao2019pcc,wang2019learning} attain the significant performance for crowd counting. Some algorithms \cite{zhang2016single, sam2017switching, idrees2018composition} design a Fully Convolutional Network (FCN) to directly generate the density map. However, the standard FCN only focuses on local features at the spatial level, which causes that the large-range contextual information can not be encoded effectively. To remedy the above problem, some methods \cite{sindagi2017cnn, sindagi2017generating, li2018csrnet, liu2018crowd} attempt to design the specific modules. However, the spatial range is not large enough so that the context is limited. In addition, traditional FCN can not encode the relation of different channels, and it is prone to predicting background as crowd region. 

\begin{figure}[t]
	\centering
	\includegraphics[width=0.9\textwidth]{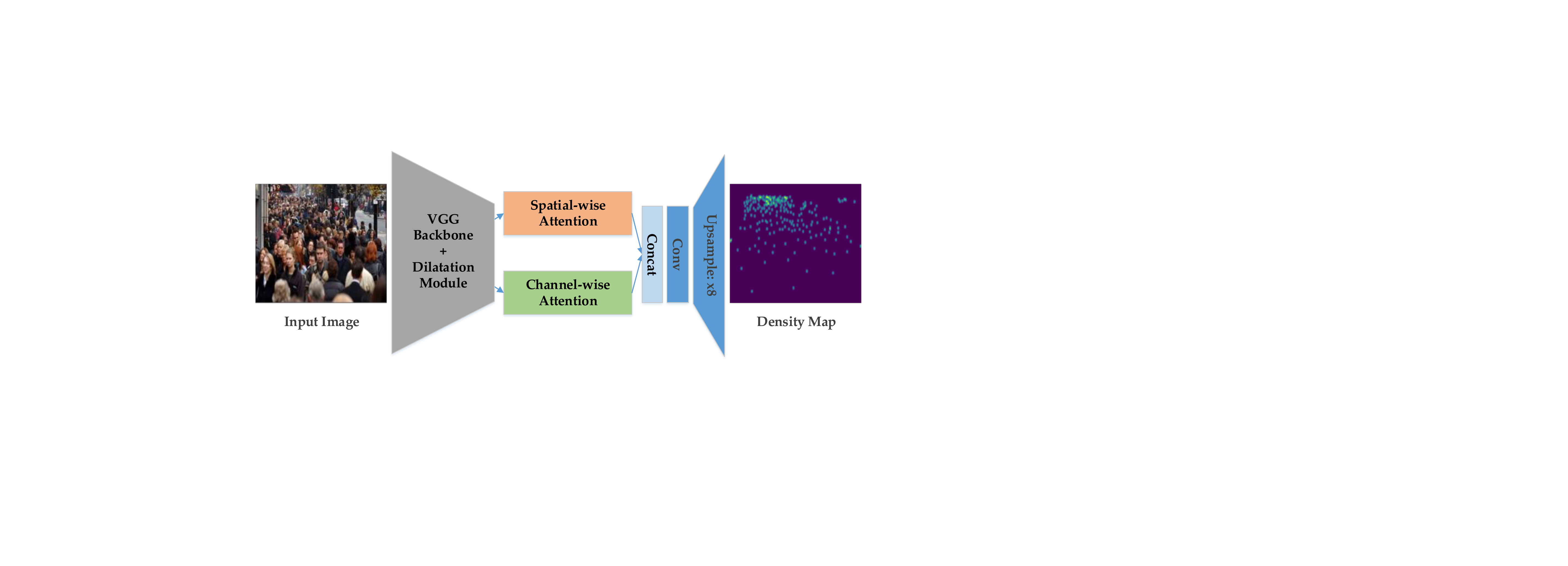}
	\caption{The flowchart of Spatial-/Channel-wise Attention Regression Networks (SCAR), which consists of two streams to encode large-range contextual information. It directly concatenates the two types of feature maps and then produces the 1-channel predicted density map via convolutional layer and up-sample operation. }\label{Fig-intro}
\end{figure}

To reduce the two problems, we propose a Spatial-/Channel-wise Attention Regression Network (SCAR) for crowd counting, which consists of Local Feature Extraction, Attention Model and Map Regressor. The architecture of the proposed networks is shown in Fig. \ref{Fig-intro}. It is a sequential pipeline, of which data is processed in turns. To be specific, Local Feature Extraction adopts the first 10 convolutional layers of VGG-16 Net \cite{simonyan2014very}, and Map Regressor consists of two convolutional layers with $1 \times 1$ kernel size. 

Here, we describe the proposed Attention Model, which is the essential module to remedy the aforementioned problems. It consists of two variants of the self-attention module \cite{vaswani2017attention,wang2018non}, which are added to the top of the traditional FCN. To be specific, the Spatial-wise Attention Model (SAM) encodes the spatial dependencies in the whole feature map, which guarantees the accurate location for the people's head. The other Channel-wise Attention Model (CAM) can handle the relations between any two-channel maps, which significantly prompts the regression performance and avoids the error estimation for backgrounds. Next, the outputs of two models will be integrated by a concatenation operation. Finally, the density map is produced by a convolutional layer according to the combined attention feature map.

In summary, the main contributions of this paper are:
\begin{enumerate}
	\item[1)] Propose Spatial-wise Attention Model (SAM) to encodes the spatial dependencies in the whole feature map, which can extract large-range contextual information.
	
	\item[2)] Present Channel-wise Attention Model (CAM) can handle the relations between any two-channel maps, which significantly prompts the regression performance and avoids the error estimation for background, especially.
	
	\item[3)] The combined model achieves the state-of-the-art on the four mainstream datasets.
\end{enumerate}

\section{Related Work}

In this section, some mainstream CNN-based crowd counting methods and important attention algorithms are briefly reviewed.

\subsection{Crowd Counting}
With the development of deep learning, many CNN-based counting models \cite{zhang2016single,sam2017switching,idrees2018composition,sindagi2017generating,Qi2017Deep,liu2018crowd,wang2019learning,ZHANG2018190} obtain good performance. Zhang \emph{et al.} \cite{zhang2016single} propose a multi-column FCN to encode the local features with the different kernel sizes. Sam \emph{et al.} \cite{sam2017switching} present a switch layer that selects specific sub-FCN to handle the image patches with different crowd density. \cite{idrees2018composition} designs a novel composition loss for FCN and obtain a good localization performance. However, the above methods only focus on the local appearance features so that they cannot encode the contextual information. Sindagi and Patel \cite{sindagi2017generating} propose a Contextual Pyramid CNN (CP-CNN) to integrate the local and global contextual information. Li \emph{et al.} \cite{li2018csrnet} embed the dilation convolution operation into the FCN to encode the contextual features. Liu \emph{et al.} \cite{liu2018crowd} propose a recurrent spatial-aware network, which can model the variations of crowd density. Gao \emph{et al.} \cite{gao2019c} propose an efficient development framework for crowd counting.

\subsection{Attention Model}

Mnih \emph{et al.} \mbox{\cite{mnih2014recurrent}} are the first to propose the visual attention via the recurrent model. After this, many researchers \mbox{\cite{sutskever2014sequence, lu2017hierarchical}} focus on encoding attention information in visual tasks. Xu \emph{et al.} \cite{xu2015show} adopt hard/soft pooling that selects/averages the most probably attentive region or the spatial features with attentive weights. Here, we list some classical attention modules in traditional CNN. Vaswani \emph{et al.} \cite{vaswani2017attention} propose a transformer architecture to dispense with recurrence and convolutions entirely, which can capture global dependencies between input and output. Chen \emph{et al.} \cite{chen2017sca} propose a sequential pipeline to encode the spatial and channel-wise Attentions in a CNN.  Non-local Neural Networks \cite{wang2018non} is proposed by Wang \emph{et al.} in 2018, which can compute the response at a position as a weighted sum of the feature maps at all positions. Woo \emph{et al.} \cite{woo2018cbam} design a Convolutional Block Attention Module (CBAM) to implement attention computation in the feed-forward convolutional neural networks.

\section{Methodology}

\subsection{Overview}

In this section, we describe the flowchart of entire networks and explain the detailed information of two attention models: SAM and CAM. 

For a crowd scene, it is firstly fed into a Local Feature Extractor, which consists of the VGG-16 backbone and dilatation module. The former is the first 10 convolutional layers of VGG-16 Net \cite{simonyan2014very}. Inspired by CSRNet, we design a dilatation module to enlarge the respective field of the extracted feature map, which outputs 1/8 size feature maps with 64-D channel. The dilatation conv's output contains more contextual information on the spatial dimension than that of the backbone. However, it still lacks more large-range spatial contextual information. In addition, it does not encode the attention features. To this end, we design two streams to respectively encode the spatial- and channel-wise attention features. To be specific, two streams are adopted the non-local module to compute large-range information. Finally, SCAR directly concatenates the two types of feature maps and then produces the 1-channel predicted density map via convolution operation. The entire architecture is described in Fig. \ref{Fig-intro}. 

During the training phase, the loss function is standard Mean Squared Error (MSE).

\begin{table}[htbp]
	\centering
	\caption{The network architecture of the proposed SCAR.}
	\begin{tabular}{p{3cm}<{\centering}|p{3cm}<{\centering}}
		\whline
		\multicolumn{2}{c}{SCAR}\\
		\whline
		\multicolumn{2}{c}{\textbf{VGG-16 backbone} }\\
		\hline
		\multicolumn{2}{c}{conv1: [k(3,3)-c64-s1-R] $\times$ 2  }\\
		\multicolumn{2}{c}{Max polling} \\
		\multicolumn{2}{c}{conv2: [k(3,3)-c128-s1-R] $\times$ 2  }\\
		\multicolumn{2}{c}{Max polling} \\
		\multicolumn{2}{c}{conv3: [k(3,3)-c128-s1-R] $\times$ 3 }\\
		\multicolumn{2}{c}{Max polling} \\
		\multicolumn{2}{c}{conv4: [k(3,3)-c512-s1-R] $\times$ 3 }\\
		\whline
		
		\multicolumn{2}{c}{\textbf{Dilation Module}}\\
		\hline
		\multicolumn{2}{c}{k(3,3)-c512-s1-d2-R}\\
		\multicolumn{2}{c}{k(3,3)-c512-s1-d2-R}\\
		\multicolumn{2}{c}{k(3,3)-c512-s1-d2-R}\\
		\multicolumn{2}{c}{k(3,3)-c256-s1-d2-R}\\
		\multicolumn{2}{c}{k(3,3)-c128-s1-d2-R}\\
		\multicolumn{2}{c}{k(3,3)-c64-s1-d2-R}\\
		\whline	
		\multicolumn{2}{c}{\textbf{Attention Model}}\\
		\hline
		SAM	&CAM\\
		\hline
		\multicolumn{2}{c}{Concat} \\
		\whline
		\multicolumn{2}{c}{\textbf{Regression Layer} } \\
		\hline
		\multicolumn{2}{c}{k(1,1)-c1-s1-R}   \\
		\multicolumn{2}{c}{Up-sample: $\times$8 }  \\
		\whline
	\end{tabular}
	\label{Table-net}
\end{table}

In order to show the detailed information, we list the hyperparameter and configuration of SCAR in Table \ref{Table-net}. In the table, ``k(3,3)-c512-s1-d2-R'' represents the convolutional operation with kernel size of $3 \times 3$, $512$ output channels, stride size of $1$ and dilation rate of $2$. The ``R'' means that the ReLU layer is added to this convolutional layer.

\subsection{Spatial-wise Attention Model}

\label{SAM}

Due to the perspective changes of crowd scenes, the global and local density distribution has a certain regularity. For the global images, the density change has a consistent gradual trend. For example, in the sample image of Fig. \ref{Fig-intro}, the density is increasing from bottom right to top left. As for the local image patches with high density, we find they have similar local patterns and texture features.

\begin{figure*}[t]
	\centering
	\includegraphics[width=0.98\textwidth]{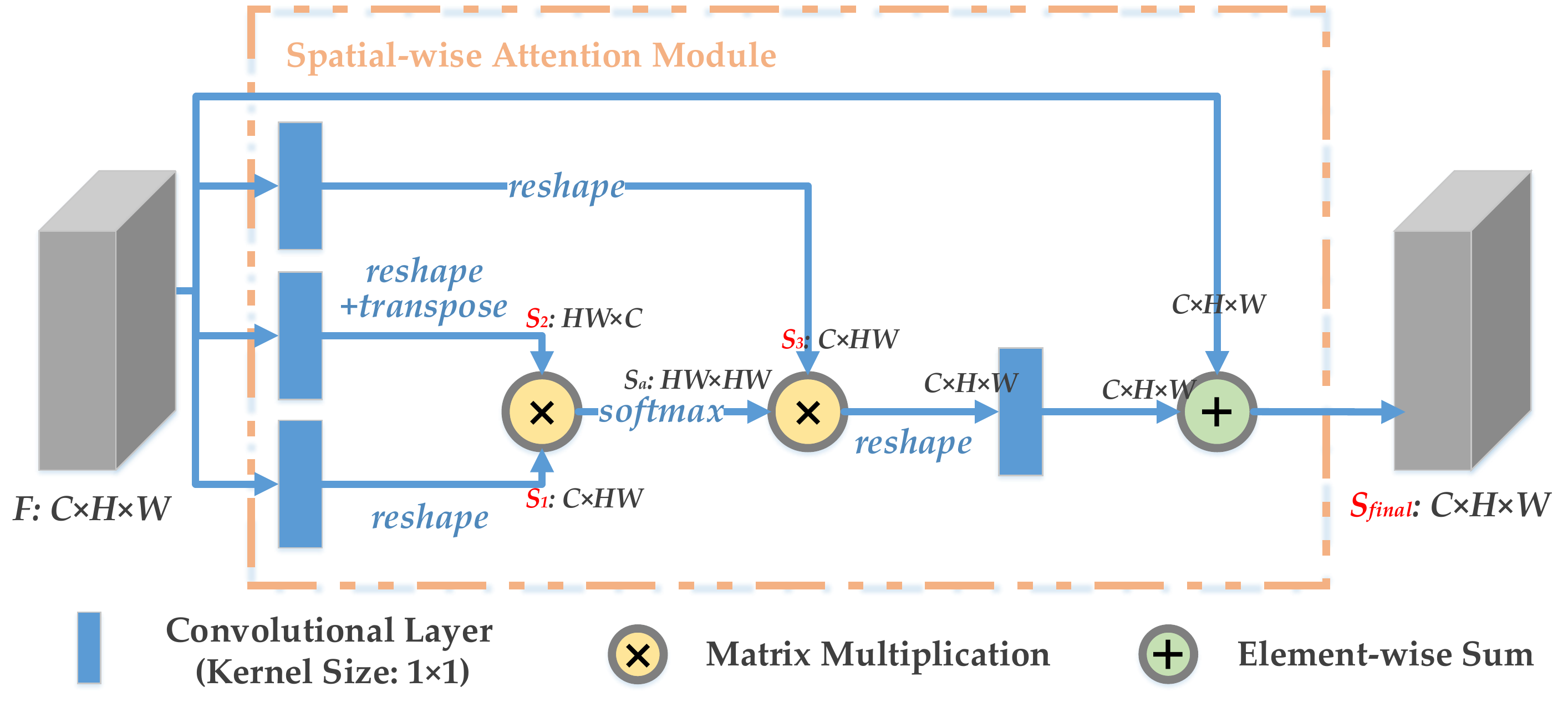}
	\caption{The detailed architectures of the Spatial-wise Attention Model (SAM) in SCAR.}\label{Fig-SAM}
\end{figure*}

In order to encode the above two observations, we design a Spatial-wise Attention Model (SAM), which can model large-range contextual information and capture the change of density distribution. The detailed architecture is described in the orange box of Fig. \ref{Fig-SAM}. For a backbone's output with the size of $C \times H \times W$, it is fed into three different $1 \times 1$ convolutional layers. Then by the reshape or transpose operations (concrete operation setting is shown in the orange box), three feature maps ${S_1}$, ${S_2}$ and ${S_3}$ are attained. For generating spatial attention map, we apply a matrix multiplication and softmax operation for ${S_1}$ and ${S_2}$. After this, we get a spatial attention map ${S_a}$ with size of $HW \times HW$. The process can be formulated as follows: 

\begin{equation}
\begin{array}{l}
\begin{aligned}
S_a^{ji} = \frac{{\exp (S_1^i \cdot S_2^j)}}{{\sum\limits_{i = 1}^{HW} {\exp (S_1^i \cdot S_2^j)} }},
\end{aligned}\label{s_a}
\end{array}
\end{equation}
where $S_a^{ji}$ represents the $i$-th position's influence on $j$-th position's. More similar feature maps of two positions have a stronger correlation of them. 

After getting ${S_a}$, we again apply a matrix multiplication between ${S_a}$ and ${S_3}$ and then reshape the output to $C \times H \times W$. For the final sum operation with $F$, we scale the output by a learnable factor. Finally, the output of SAM is defined as below:

\begin{equation}
\begin{array}{l}
\begin{aligned}
S_{final}^j = \lambda \sum\limits_{i = 1}^{HW} {\left( {S_a^{ji} \cdot S_3^i} \right)}  + {F^j},
\end{aligned}\label{s_final}
\end{array}
\end{equation}
where $\lambda$ is a learnable parameter. In the practice, we exploit a convolutional layer with kernel size $1 \times 1$ to learn the $\lambda$. 

From the entire detailed description of SAM, the final output feature map $S_{final}$ is weighted sum of attention map and original local features map, which contains global contextual features and self-attention information. Thus, SAM effectively tackles our two observations mentioned at the beginning of this section.

\subsection{Channel-wise Attention Model}

In the last section, SAM attempts encode large-range dependencies on a spatial dimension, which is effective for the performance of density location. In order to prompt class-specific response, we design a similar structure to SAM to learn dependencies on channel dimension, which is called as ``Channel-wise Attention Model''. In the field of crowd counting, the class-specific response consists of two types: foreground (head region) and background (other region). As for highly congested crowd scenes, the foreground's textures are very similar to that of some background region (tree, building and so on). Embedding CAM can effectively remedy the above estimation errors. 

\begin{figure*}[t]
	\centering
	\includegraphics[width=0.98\textwidth]{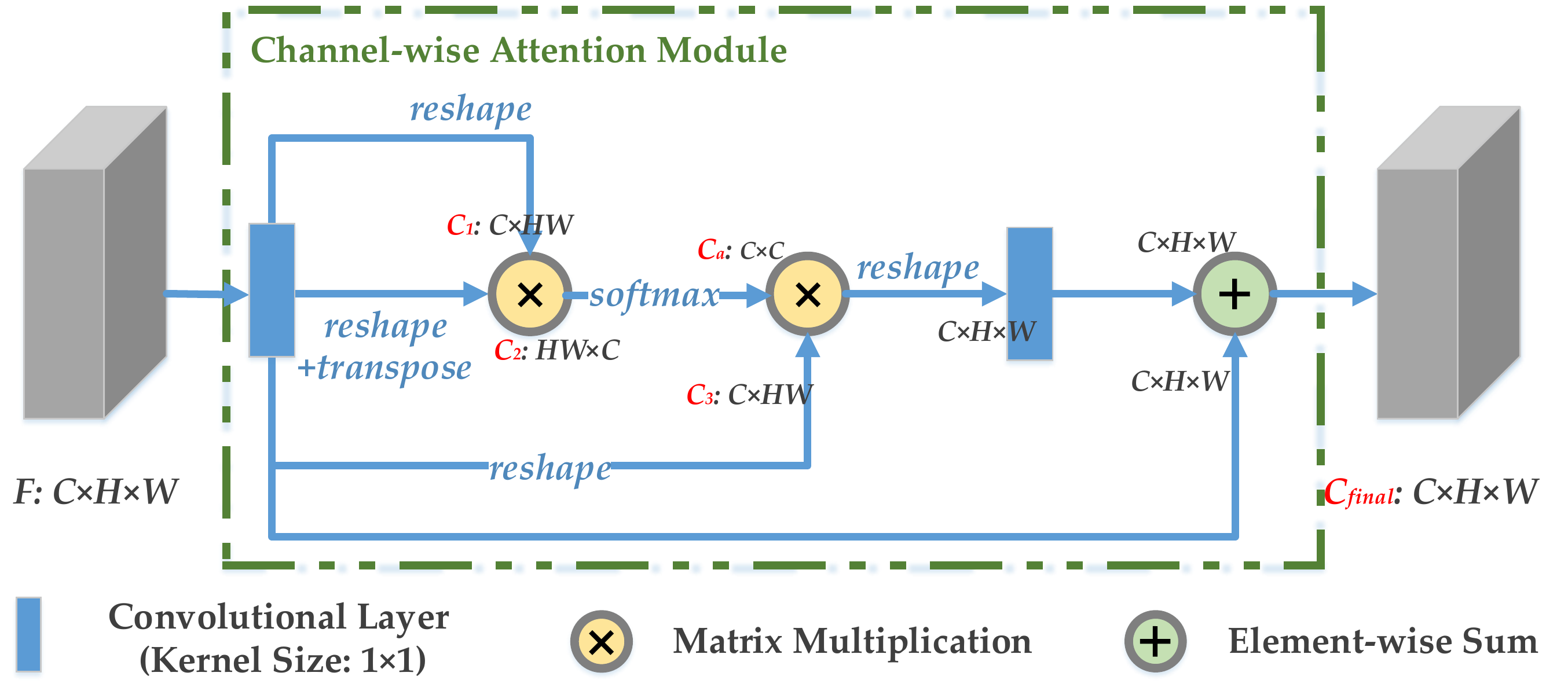}
	\caption{The detailed architectures of the Channel-wise Attention Model (CAM) in SCAR.}\label{Fig-CAM}
\end{figure*}

The concrete architecture of CAM is demonstrated in the green box of Fig. \ref{Fig-CAM}. Compared with SAM, CAM has two differences as below:
\begin{enumerate}
	\item[1)] CAM has only one $1 \times 1$ convolutional layer to tackle obtained feature map from the backbone, but SAM has threes.
	\item[2)] The sizes of intermediate feature maps are different, and the detailed values are shown in Fig. \mbox{\ref{Fig-CAM}}.
\end{enumerate}

Similarly, the main operations are the same as SAM. To be specific, ${C_a}$ with size of $C \times C$ is defined as:
\begin{equation}
\begin{array}{l}
\begin{aligned}
C_a^{ji} = \frac{{\exp (C_1^i \cdot C_2^j)}}{{\sum\limits_{i = 1}^C {\exp (C_1^i \cdot C_2^j)} }},
\end{aligned}\label{c_a}
\end{array}
\end{equation}
where $C_a^{ji}$ denotes the $i$-th channel's influence on $j$-th channel's. And $C_{final}$ with a size of $C \times H \times W$ is computed as:
\begin{equation}
\begin{array}{l}
\begin{aligned}
C_{final}^j = \mu \sum\limits_{i = 1}^C {\left( {C_a^{ji} \cdot C_3^i} \right)}  + {F^j},
\end{aligned}\label{c_final}
\end{array}
\end{equation}
where $\mu$ is a learnable parameter. In practice, we exploit a convolutional layer with kernel size $1 \times 1$ to learn the $\mu$, which is same as the $\lambda$ in Sec. \mbox{\ref{SAM}}.

\section{Experiments}

\subsection{Evaluation Metrics}
In this paper, we evaluate the methods from two perspectives: the counting performance and the quality of density maps. To be specific, for the former, the Mean Absolute Error (MAE) and Mean Squared Error (MSE) are introduced into each model or algorithm, which are defined as follows: 
\begin{equation}
\begin{array}{l}
\begin{aligned}
\boldsymbol{MAE} = \frac{1}{N}\sum\limits_{i = 1}^N {\left| {{y_i} - {{\hat y}_i}} \right|}, 
\end{aligned}\label{MAE}
\end{array}
\end{equation}

\begin{equation}
\begin{array}{l}
\begin{aligned}
\boldsymbol{MSE} = \sqrt {\frac{1}{N}\sum\limits_{i = 1}^N {{{\left| {{y_i} - {{\hat y}_i}} \right|}^2}} } ,
\end{aligned}\label{MSE}
\end{array}
\end{equation}
where $N$ is the number of images in testing set, ${{y_i}}$ is the ground truth of people number and ${{{\hat y}_i}}$ is the estimated count value for the $i$th testing image. 

For further evaluating the quality of density maps, we exploit PSNR (Peak Signal-to-Noise Ratio) and SSIM (Structural Similarity in Image \cite{wang2004image}), which are the full reference metrics.

\subsection{Implementation Details}
In our experiments, all images are resized to $576 \times 768$, and the density maps are generated under the same size. The learning rates of the entire networks are initialized at ${10^{ - 5}}$ and reduced to $0.995$ times every epoch. The batch size is set as $4$ on each GPU. The Adam algorithm is exploited to optimize the proposed networks and obtain the best results after 400 epochs. Take the experiment on Shanghai Tech Part B Dataset as an example, the entire training process spends 4 hours on  two paralleled GPUs. In SAM and CAM, the channel number C is set as 64, which is the channel number of Dilatation Module's outputs.

All experimental training and evaluation are performed on NVIDIA GTX 1080Ti GPU using PyTorch framework \cite{paszke2017pytorch}.

\subsection{Performance on Shanghai Tech Dataset}

ShanghaiTech dataset \mbox{\cite{zhang2016single}} is a real-world crowd counting dataset, which is collected by the researchers of ShanghaiTech University. It consists of two parts: A and B. To be specific, Part A is collected from a photo-sharing website (https://www.flickr.com/), of which images have different resolution. Part B contains 400 training and 316 testing images. of which crowd scenes are captured from surveillance cameras, which are installed on the walking streets in Shanghai, China.

\begin{table}[htbp]
	
	\centering
	\caption{Estimation errors on ShanghaiTech dataset.}
	
	\begin{tabular}{c|cc|cc}
		\whline
		\multirow{2}{*}{Method} &	\multicolumn{2}{c|}{Part A} &			\multicolumn{2}{c}{Part B} 	\\
		\cline{2-5} 
		
		&MAE 		&MSE			&MAE			&MSE 		 		\\
		\whline
		Zhang \emph{et al.} \cite{zhang2015cross} &181.8 	&277.7	&32.0 &49.8			\\
		\hline
		MCNN \cite{zhang2016single}&110.2& 173.2& 26.4 &41.3 \\
		\hline
		Switching-CNN \cite{sam2017switching} &90.4& 135.0& 21.6 &33.4 \\
		\hline
		CP-CNN \cite{sindagi2017generating} &73.6  &\textbf{106.4}  &20.1 & 30.1  \\
		\hline
		DR-ResNet \cite{ding2018deeply}  &86.3 &124.2 &14.5 &21.0 \\
		\hline
		CSRNet \cite{li2018csrnet}  &68.2 &115.0 &10.6 &16.0  \\
		\hline
		ic-CNN \cite{ranjan2018iterative}  &68.5 &116.2 &10.7 &16.0 \\
		\whline
		SCAR (ours)  &\textbf{66.3} 	&114.1	&\textbf{9.5}  &\textbf{15.2} \\
		\whline
		
	\end{tabular}\label{shanghai}
\end{table}

Table \ref{shanghai} reports the results of some mainstream methods on ShanghaiTech dataset. From it, the proposed SCAR wins three first places and one second place. Compared with Switching-CNN \cite{sam2017switching} and CSRNet \cite{li2018csrnet} (the both also adopt VGG-16 as the pre-trained model), our model is the best from the overall performance on the two datasets. 

\begin{figure*}[htbp]
	\centering
	\includegraphics[width=1\textwidth]{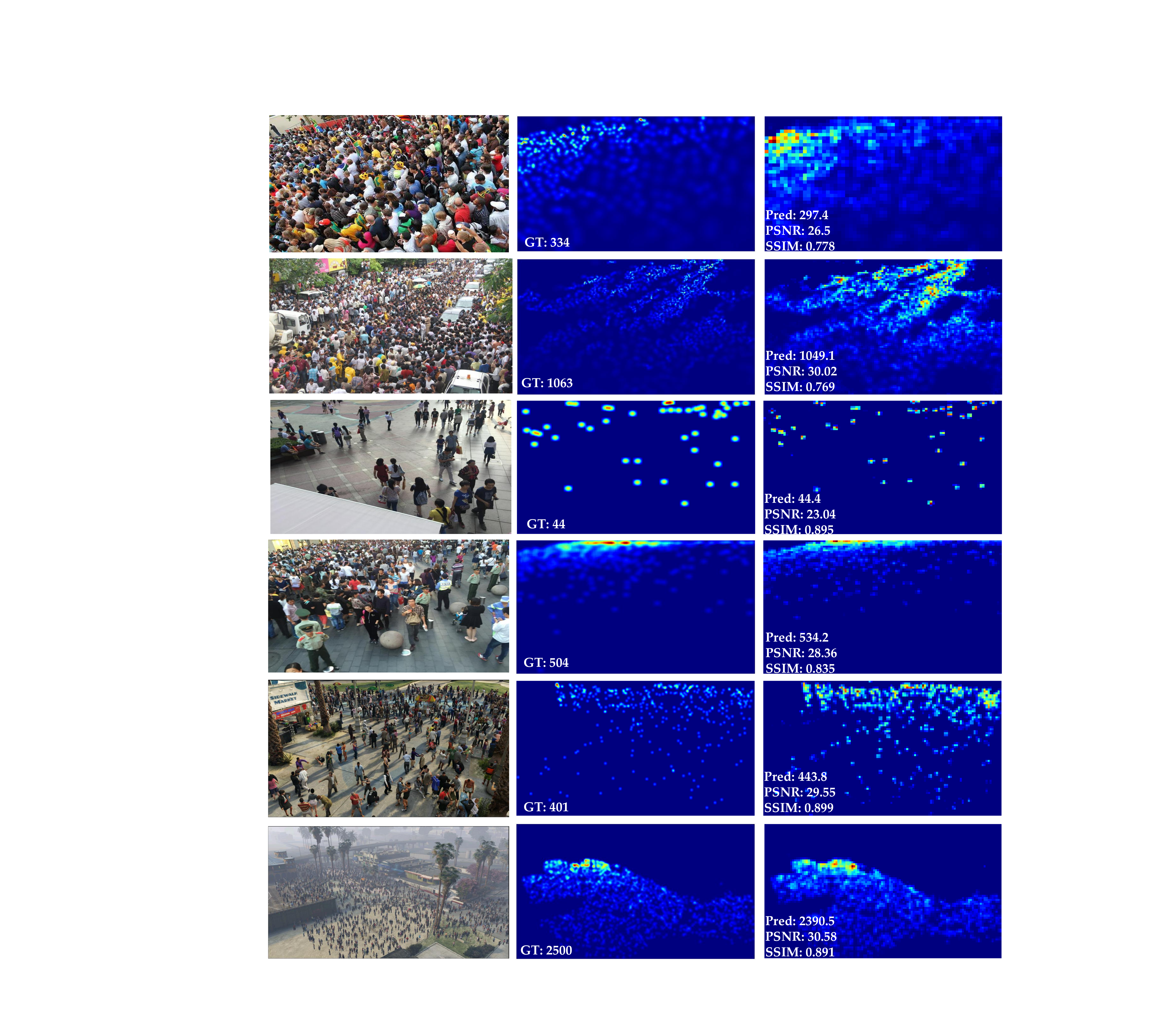}
	\caption{Exemplar results of the full model on Shanghai Tech Part A, B and GCC Dataset. Column 1: input image, Column 2: groundtruth, Column 3: predicted density map by SCAR. ``GT '' and ``Pred'' denote the ground truth count and the estimation count, respectively.}\label{Fig_show}
\end{figure*}

In order to intuitively show the performance of SCAR, Fig. \mbox{\ref{Fig_show}} illustrates the six groups of visualization results on Shanghai Part A, B and GCC Dataset. The first, second and third column demonstrates the original images, groundtruth and predicted density maps of SCAR, respectively. From it, we find the predict maps can show the densities of different regions and the estimation counting numbers are very close to the label counting numbers.

\subsection{Performance on UCF\_CC\_50}

UCF\_CC\_50 dataset is extremely congested crowd counting dataset, which is released by Idrees \emph{et al.} \cite{idrees2013multi}. It only contains 50 images without partition for training and testing. Thus, we adopt the 5-fold cross-validation protocol to evaluate SCAR with other methods. Table \ref{UCF} lists the recent popular methods' results on this dataset. From the table, we achieve the best MAE of 259.0 and the third-place MSE of 374.0. 

\begin{table}[htbp]
	\centering
	\caption{Estimation errors on UCF\_CC\_50 dataset.}
	
	\begin{tabular}{c|cc}
		\whline
		Methods 				 &MAE 		&MSE	 		\\
		\whline
		Idrees \emph{et al.} \cite{idrees2013multi} &419.5& 541.6		\\
		\hline
		Zhang \emph{et al.} \cite{zhang2015cross} &467.0& 498.5	\\
		\hline
		MCNN \cite{zhang2016single}&377.6& 509.1 \\
		\hline			
		Switching-CNN \cite{sam2017switching} &318.1 &439.2 \\
		\hline
		CP-CNN \cite{sindagi2017generating} &295.8&\textbf{320.9}   \\
		\hline
		DR-ResNet \cite{ding2018deeply}  &307.4 & 421.6 \\
		\hline
		CSRNet \cite{li2018csrnet}  &266.1 & 397.5  \\
		\hline
		ic-CNN \cite{ranjan2018iterative}  &260.9 & 365.5  \\
		\whline
		SCAR (ours)  &\textbf{259.0} 	&374.0  \\
		\whline
		
	\end{tabular}\label{UCF}
\end{table}

\subsection{Performance on GCC}

GTA V Crowd Counting Dataset (GCC) \cite{wang2019learning} is a large-scale synthetic dataset based on an electronic game, which consists of  15,212 crowd images. GCC provides three evaluation strategies (random splitting, cross-camera, and cross-location evaluation) to show the capacity from different angles. In this section, we implement SCAR on GCC dataset and the results are reported in Table \ref{Table-gcc}. At the same time, the results of four classical models (MCNN \cite{zhang2016single}, CSRNet\cite{li2018csrnet}, FCN \cite{wang2019learning} and SFCN \cite{wang2019learning}) are listed in Table \ref{Table-gcc}. 

\begin{table}[htbp]
	\centering
	\caption{The results of our proposed SCAR and the four classic methods on GCC dataset.}
	\begin{tabular}{c|cc|cc}
		\multicolumn{5}{c}{\small{Performance of random splitting}}\\
		\whline
		Method	  &MAE &MSE &PSNR &SSIM  \\
		\whline
		MCNN \cite{zhang2016single}   &100.9 &217.6 &24.00 &0.838 \\
		\hline
		CSRNet\cite{li2018csrnet}  &38.2 &87.6 &29.52 &0.829 \\
		\hline
		FCN \cite{wang2019learning}&42.3 &98.7 &30.10 &0.889\\
		\hline	
		SFCN \cite{wang2019learning}&36.2 &81.1 &30.21 &0.904\\
		\whline
		SCAR &\textbf{31.7} &\textbf{76.8} &\textbf{30.56} &\textbf{0.921}\\
		\whline		
	\end{tabular}
	
	\begin{tabular}{c|cc|cc}
		\multicolumn{5}{c}{\small{Performance of cross-camera splitting}}\\
		\whline
		Method	  &MAE &MSE &PSNR &SSIM  \\
		\whline
		MCNN \cite{zhang2016single}   &110.0 &221.5 &23.81 &0.842 \\
		\hline
		CSRNet\cite{li2018csrnet}  &61.1 &134.9 &29.03 &0.826 \\
		\hline
		FCN \cite{wang2019learning}&61.5 &156.6 &28.92 &0.874\\
		\hline	
		SFCN \cite{wang2019learning}&56.0 &\textbf{129.7} &\textbf{29.17} &0.889\\
		\whline
		SCAR &\textbf{55.8} &135.3 &28.84 &\textbf{0.894}\\
		\whline
	\end{tabular}
	\begin{tabular}{c|cc|cc}
		\multicolumn{5}{c}{\small{Performance of cross-location splitting}}\\
		\whline
		Method	  &MAE &MSE &PSNR &SSIM  \\
		\whline
		MCNN \cite{zhang2016single}   &154.8 &340.7 &24.05 &0.857 \\
		\hline
		CSRNet\cite{li2018csrnet}  &92.2 &220.1 &28.75 &0.842 \\
		\hline
		FCN \cite{wang2019learning}&97.5 &226.8 &29.33 &0.866\\
		\hline	
		SFCN \cite{wang2019learning}&89.3 &\textbf{216.8} &29.50 &0.906\\
		\whline
		SFCN &\textbf{87.2} &220.7 &\textbf{29.74} &\textbf{0.929}\\
		\whline
	\end{tabular}
	\label{Table-gcc}
\end{table}

From the table, we find our SCAR attains the best place of MAE (random splitting/cross-camera/cross-location evaluation: \textbf{31.7/55.8/87.2}) compared with other mainstream methods. To be specific, CSRNet\cite{li2018csrnet}, FCN \cite{wang2019learning},  SFCN \cite{wang2019learning} and the proposed SCAR adopt the same backbone, the first 10 layers of VGG-16 Net \cite{simonyan2014very}. Compared with them, our SCAR achieves the 8 best places from the 12 metrics, which shows that the proposed attention model can effectively prompt the estimation performance for crowd density.

\section{Discussion and Analysis}

\subsection{Ablation Study on ShanghaiTech Part B}
\label{ablation}
For showing the effect of each module (SAM and CAM), we conduct the ablation experiments on the ShanghaiTech Part B Dataset \cite{zhang2016single}. Table \ref{Table-ablation} demonstrates the performance of models with four different settings:

\textbf{FCN}: the baseline of this paper. It is the combination of a single-column VGG-16 FCN and the dilation conv in CSRNet \cite{li2018csrnet};

\textbf{FCN+SAM}: SAM is added to the top of FCN;

\textbf{FCN+CAM}: CAM is added to the top of FCN;

\textbf{SCAR}: the full model is proposed by ours, which consists of FCN, SAM and CAM.

\begin{table}[htbp]
	
	\centering
	\caption{Estimation errors and density map quality for different models of the proposed method on ShanghaiTech Part B dataset.}
	
	\begin{tabular}{c|cc|cc}
		\whline
		Methods 				 &MAE 		&MSE	 & PSNR	& SSIM	\\
		\whline
		FCN            &13.2 &21.0 &21.36 &0.787 	\\
		\hline
		FCN+SAM     &11.0   &18.8 &23.01 &0.881	\\
		FCN+CAM     &11.5   &19.3 &22.45 &0.875	\\
		SCAR     &\textbf{9.5}  &\textbf{15.2}	&\textbf{24.03} &\textbf{0.912}\\
		\whline
		
	\end{tabular}\label{Table-ablation}
\end{table}

From the table, we find that SAM achieves a better performance than CAM: \textbf{11.0/18.8} of MAE/MSE v.s. 11.5/19.3 of MAE/MSE. Meanwhile, FCN+SAM produces more high-quality density maps than FCN+CAM, namely \textbf{23.01/0.881} of PSNR/SSIM v.s. 22.45/0.875 of PSNR/SSIM. When simultaneously embedding SAM and CAM into the networks, the proposed model attains the best performance regardless of the counting results or density maps' quality. 

\subsection{Comparison of Density Map Quality}

The evaluation of density map quality is an emerging criterion in this field, which is adopted by \cite{li2018csrnet}. In \cite{li2018csrnet}, the authors use two criteria (PSNR and SSIM) to evaluate the quality of density maps. Here, we also compared the above metrics with MCNN \cite{zhang2016single}, CP-CNN \cite{sindagi2017generating} and CSRNet \cite{li2018csrnet}. Table \ref{quality} shows that our SCAR outperforms other three methods. Particularly, we obtains the 23.93 of PSNR and 0.81 of SSIM on Shanghai Tech Part A dataset.

\begin{table}[htbp]	
	\centering
	\caption{Density map quality on ShanghaiTech Part A.}
	
	\begin{tabular}{c|cc}
		\whline
		Methods 				 &PSNR 		&SSIM	 		\\
		\whline
		MCNN \cite{zhang2016single} &21.4& 0.52 \\
		\hline			
		CP-CNN \cite{sindagi2017generating} &21.72&0.72   \\
		\hline
		CSRNet \cite{li2018csrnet}  &23.79 &0.76   \\
		\hline
		SCAR (ours)  &\textbf{23.93} &\textbf{0.81}  \\
		\whline
		
	\end{tabular}\label{quality}
\end{table}

\subsection{Analysis of Attention Feature Fusion}

In this paper, for two types of attention features, namely outputs of SAM and CAM, we adopt concatenation operation to fuse them and then produce 1-channel density map. In addition to the concatenation operation, the element-wise sum is also a potential operation to integrate different features. In this section, we compare the performance of different fusion strategies of two types of attention features. To be specific, we conduct a comparison experiment on Shanghai Tech Part B dataset, and the experimental results are reported in Table \mbox{\ref{Table-operation}}. 

\begin{table}[htbp]
	
	\centering
	\caption{Estimation errors of different different fusion strategies on ShanghaiTech Part B dataset.}
	
	\begin{tabular}{c|cc|cc}
		\whline
		Fusion Strategy		 &MAE 		&MSE	 & PSNR	& SSIM	\\
		\whline
		Element-wise sum  &9.9   &16.7 & 23.91 & 0.898\\
		Concatenation     &\textbf{9.5}  &\textbf{15.2}	&\textbf{24.03} &\textbf{0.912}\\
		\whline
		
	\end{tabular}\label{Table-operation}
\end{table}

From the table, we find the concatenation operation obtains lower estimation errors compared with the element-wise sum operation, which shows the feature fusion effect of the former is better than that of the latter.


\subsection{Attention Map Visualization}

\begin{figure*}[htbp]
	\centering
	\includegraphics[width=1\textwidth]{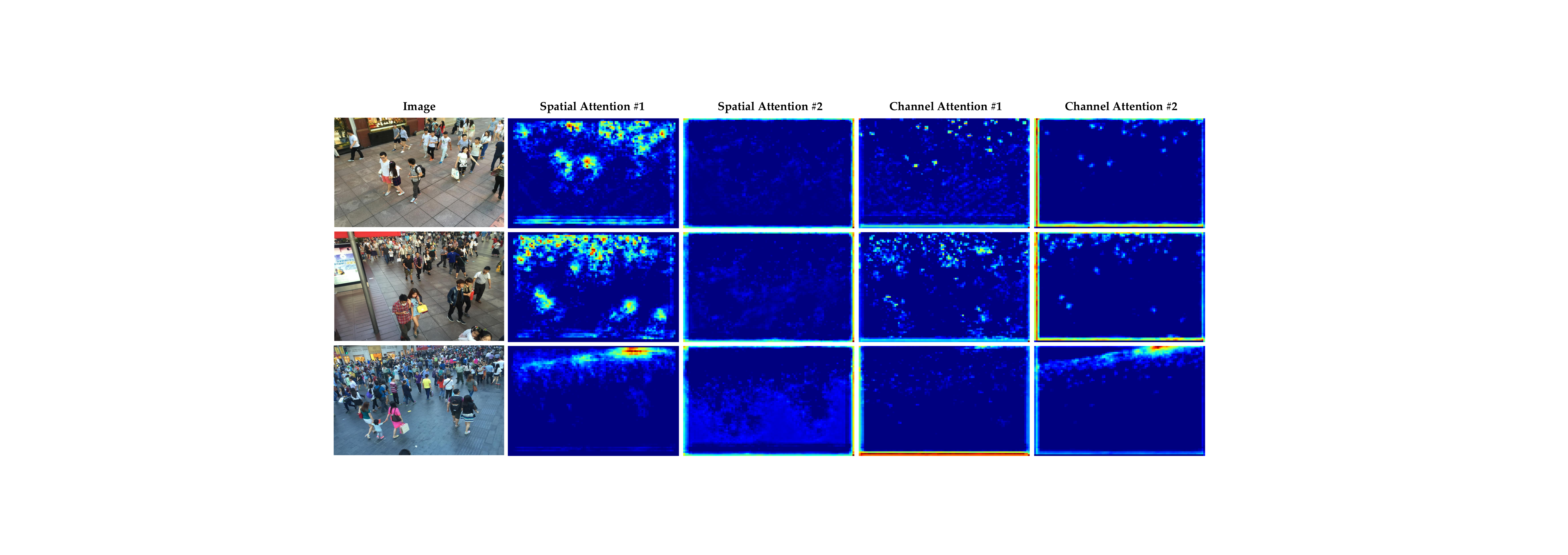}
	\caption{Exemplar results of the different attention maps. Column 1: input image, Column 2 and 3: Spatial-wise attention maps, Column 3 and 4: Channel-wise attention maps.}\label{Fig_att}
\end{figure*}

In proposed SCAR, we design two types of attention maps, namely spatial-wise and channel-wise maps. The former focuses on encoding the pixel-wise context of the entire image. The latter attempts to extract more discriminative features among different channels, which aids model to pay attention to the head region. Here, we analyze the effect of different attention maps by visualizing them. Fig. \mbox{\ref{Fig_att}} shows the results of attention maps in some typical crowd scenes. To be specific, we select \#1 and \#2 maps from 64 channels in two types attention masks. 

From the figure, we find that spatial attention maps can capture the large-range context information. The \#1 is sensitive to people region the \#2 can effectively segment the background. For the channel attention maps (\#1 and \#2), they can accurately locate the head position. The above visualization results intuitively show different effects of SAM and CAM in the proposed SCAR. Finally, the proposed SCAR encodes context and location information to predict the density map.

\section{Conclusion}

This paper proposes a Spatial-/Channel-wise Attention Regression Network (SCAR) to generate high-quality density map and estimate the number of people in crowd scenes. To be specific, spatial-wise attention model and Channel-wise attention model are parallel architecture on the top of VGG-16 backbone network. SCAR is very flexible, of which attention modules can be embedded into any CNN to encode large-range contextual information. Thus, we believe that it is applied to other pixel-wise tasks, such as saliency detection, image segmentation and so on. In the future, we will verify the guess on the above applications. 


\bibliographystyle{elsarticle-num} 
\bibliography{strings}

\begin{thebibliography}{10}
\expandafter\ifx\csname url\endcsname\relax
  \def\url#1{\texttt{#1}}\fi
\expandafter\ifx\csname urlprefix\endcsname\relax\def\urlprefix{URL }\fi
\expandafter\ifx\csname href\endcsname\relax
  \def\href#1#2{#2} \def\path#1{#1}\fi

\bibitem{ZITOUNI2016139}
M.~S. Zitouni, H.~Bhaskar, J.~Dias, M.~Al-Mualla, Advances and trends in visual
  crowd analysis: A systematic survey and evaluation of crowd modelling
  techniques, Neurocomputing 186 (2016) 139 -- 159.

\bibitem{LU2017213}
W.~Lu, X.~Wei, W.~Xing, W.~Liu, Trajectory-based motion pattern analysis of
  crowds, Neurocomputing 247 (2017) 213 -- 223.

\bibitem{WEI2019}
X.~Wei, J.~Du, Z.~Xue, M.~Liang, Y.~Geng, X.~Xu, J.~Lee, A very deep two-stream
  network for crowd type recognition, Neurocomputing.

\bibitem{CHEUNG20191}
E.~Cheung, A.~Wong, A.~Bera, X.~Wang, D.~Manocha, Lcrowdv: Generating labeled
  videos for pedestrian detectors training and crowd behavior learning,
  Neurocomputing 337 (2019) 1 -- 14.

\bibitem{ZHANG2015151}
Z.~Zhang, M.~Wang, X.~Geng, Crowd counting in public video surveillance by
  label distribution learning, Neurocomputing 166 (2015) 151 -- 163.

\bibitem{li2018csrnet}
Y.~Li, X.~Zhang, D.~Chen, Csrnet: Dilated convolutional neural networks for
  understanding the highly congested scenes, in: Proceedings of the IEEE
  Conference on Computer Vision and Pattern Recognition, 2018, pp. 1091--1100.

\bibitem{WANG2019360}
L.~Wang, B.~Yin, X.~Tang, Y.~Li, Removing background interference for crowd
  counting via de-background detail convolutional network, Neurocomputing 332
  (2019) 360 -- 371.

\bibitem{MA201991}
J.~Ma, Y.~Dai, Y.-P. Tan, Atrous convolutions spatial pyramid network for crowd
  counting and density estimation, Neurocomputing 350 (2019) 91 -- 101.

\bibitem{lu2017hybrid}
X.~Lu, W.~Zhang, X.~Li, A hybrid sparsity and distance-based discrimination
  detector for hyperspectral images, IEEE Transactions on Geoscience and Remote
  Sensing 56~(3) (2017) 1704--1717.

\bibitem{lu2017exploring}
X.~Lu, B.~Wang, X.~Zheng, X.~Li, Exploring models and data for remote sensing
  image caption generation, IEEE Transactions on Geoscience and Remote Sensing
  56~(4) (2017) 2183--2195.

\bibitem{zhao2017hierarchical}
B.~Zhao, X.~Li, X.~Lu, Hierarchical recurrent neural network for video
  summarization, in: Proceedings of the 25th ACM international conference on
  Multimedia, ACM, 2017, pp. 863--871.

\bibitem{zhao2019cam}
B.~Zhao, X.~Li, X.~Lu, Cam-rnn: Co-attention model based rnn for video
  captioning, IEEE Transactions on Image Processing.

\bibitem{8766103}
J.~{Gao}, Q.~{Wang}, Y.~{Yuan}, Convolutional regression network for
  multi-oriented text detection, IEEE Access 7 (2019) 96424--96433.

\bibitem{8693661}
Q.~{Wang}, J.~{Gao}, X.~{Li}, Weakly supervised adversarial domain adaptation
  for semantic segmentation in urban scenes, IEEE Transactions on Image
  Processing 28~(9) (2019) 4376--4386.

\bibitem{gao2019pcc}
J.~Gao, Q.~Wang, X.~Li, Pcc net: Perspective crowd counting via spatial
  convolutional network, IEEE Transactions on Circuits and Systems for Video
  Technology.

\bibitem{wang2019learning}
Q.~Wang, J.~Gao, W.~Lin, Y.~Yuan, Learning from synthetic data for crowd
  counting in the wild, in: Proceedings of IEEE Conference on Computer Vision
  and Pattern Recognition (CVPR), 2019.

\bibitem{zhang2016single}
Y.~Zhang, D.~Zhou, S.~Chen, S.~Gao, Y.~Ma, Single-image crowd counting via
  multi-column convolutional neural network, in: Proceedings of the IEEE
  conference on computer vision and pattern recognition, 2016, pp. 589--597.

\bibitem{sam2017switching}
D.~B. Sam, S.~Surya, R.~V. Babu, Switching convolutional neural network for
  crowd counting, in: Proceedings of the IEEE Conference on Computer Vision and
  Pattern Recognition, Vol.~1, 2017, p.~6.

\bibitem{idrees2018composition}
H.~Idrees, M.~Tayyab, K.~Athrey, D.~Zhang, S.~Al-Maadeed, N.~Rajpoot, M.~Shah,
  Composition loss for counting, density map estimation and localization in
  dense crowds, arXiv preprint arXiv:1808.01050.

\bibitem{sindagi2017cnn}
V.~A. Sindagi, V.~M. Patel, Cnn-based cascaded multi-task learning of
  high-level prior and density estimation for crowd counting, in: Advanced
  Video and Signal Based Surveillance (AVSS), 2017 14th IEEE International
  Conference on, IEEE, 2017, pp. 1--6.

\bibitem{sindagi2017generating}
V.~A. Sindagi, V.~M. Patel, Generating high-quality crowd density maps using
  contextual pyramid cnns, in: 2017 IEEE International Conference on Computer
  Vision (ICCV), IEEE, 2017, pp. 1879--1888.

\bibitem{liu2018crowd}
L.~Liu, H.~Wang, G.~Li, W.~Ouyang, L.~Lin, Crowd counting using deep recurrent
  spatial-aware network, arXiv preprint arXiv:1807.00601.

\bibitem{simonyan2014very}
K.~Simonyan, A.~Zisserman, Very deep convolutional networks for large-scale
  image recognition, arXiv preprint arXiv:1409.1556.

\bibitem{vaswani2017attention}
A.~Vaswani, N.~Shazeer, N.~Parmar, J.~Uszkoreit, L.~Jones, A.~N. Gomez,
  {\L}.~Kaiser, I.~Polosukhin, Attention is all you need, in: Advances in
  Neural Information Processing Systems, 2017, pp. 5998--6008.

\bibitem{wang2018non}
X.~Wang, R.~Girshick, A.~Gupta, K.~He, Non-local neural networks, in: The IEEE
  Conference on Computer Vision and Pattern Recognition (CVPR), 2018.

\bibitem{Qi2017Deep}
W.~Qi, W.~Jia, Y.~Yuan, Deep metric learning for crowdedness regression, IEEE
  Transactions on Circuits and Systems for Video Technology PP~(99) (2017)
  1--1.

\bibitem{ZHANG2018190}
Y.~Zhang, F.~Chang, M.~Wang, F.~Zhang, C.~Han, Auxiliary learning for crowd
  counting via count-net, Neurocomputing 273 (2018) 190 -- 198.

\bibitem{gao2019c}
J.~Gao, W.~Lin, B.~Zhao, D.~Wang, C.~Gao, J.~Wen, C$^3$ framework: An
  open-source pytorch code for crowd counting, arXiv preprint arXiv:1907.02724.

\bibitem{mnih2014recurrent}
V.~Mnih, N.~Heess, A.~Graves, et~al., Recurrent models of visual attention, in:
  Advances in neural information processing systems, 2014, pp. 2204--2212.

\bibitem{sutskever2014sequence}
I.~Sutskever, O.~Vinyals, Q.~V. Le, Sequence to sequence learning with neural
  networks, in: Advances in neural information processing systems, 2014, pp.
  3104--3112.

\bibitem{lu2017hierarchical}
X.~Lu, Y.~Chen, X.~Li, Hierarchical recurrent neural hashing for image
  retrieval with hierarchical convolutional features, IEEE Transactions on
  Image Processing 27~(1) (2017) 106--120.

\bibitem{xu2015show}
K.~Xu, J.~Ba, R.~Kiros, K.~Cho, A.~Courville, R.~Salakhutdinov, R.~Zemel,
  Y.~Bengio, Show, attend and tell: Neural image caption generation with visual
  attention, arXiv preprint arXiv:1502.03044.

\bibitem{chen2017sca}
L.~Chen, H.~Zhang, J.~Xiao, L.~Nie, J.~Shao, W.~Liu, T.-S. Chua, Sca-cnn:
  Spatial and channel-wise attention in convolutional networks for image
  captioning, in: Proceedings of the IEEE conference on computer vision and
  pattern recognition, 2017, pp. 5659--5667.

\bibitem{woo2018cbam}
S.~Woo, J.~Park, J.-Y. Lee, I.~So~Kweon, Cbam: Convolutional block attention
  module, in: Proceedings of the European Conference on Computer Vision (ECCV),
  2018, pp. 3--19.

\bibitem{wang2004image}
Z.~Wang, A.~C. Bovik, H.~R. Sheikh, E.~P. Simoncelli, Image quality assessment:
  from error visibility to structural similarity, IEEE transactions on image
  processing 13~(4) (2004) 600--612.

\bibitem{paszke2017pytorch}
A.~Paszke, S.~Gross, S.~Chintala, G.~Chanan, Pytorch: Tensors and dynamic
  neural networks in python with strong gpu acceleration (2017).

\bibitem{zhang2015cross}
C.~Zhang, H.~Li, X.~Wang, X.~Yang, Cross-scene crowd counting via deep
  convolutional neural networks, in: Computer Vision and Pattern Recognition
  (CVPR), 2015 IEEE Conference on, IEEE, 2015, pp. 833--841.

\bibitem{ding2018deeply}
X.~Ding, Z.~Lin, F.~He, Y.~Wang, Y.~Huang, A deeply-recursive convolutional
  network for crowd counting, arXiv preprint arXiv:1805.05633.

\bibitem{ranjan2018iterative}
V.~Ranjan, H.~Le, M.~Hoai, Iterative crowd counting, arXiv preprint
  arXiv:1807.09959.

\bibitem{idrees2013multi}
H.~Idrees, I.~Saleemi, C.~Seibert, M.~Shah, Multi-source multi-scale counting
  in extremely dense crowd images, in: Computer Vision and Pattern Recognition
  (CVPR), 2013 IEEE Conference on, IEEE, 2013, pp. 2547--2554.

\end{thebibliography}


\end{document}